\documentclass[letterpaper]{article} 
\usepackage{aaai24}  
\usepackage{times}  
\usepackage{helvet}  
\usepackage{courier}  
\usepackage[hyphens]{url}  
\usepackage{graphicx} 
\usepackage{natbib}  
\usepackage{caption} 
\usepackage{algorithm}
\usepackage{algorithmic}

\usepackage{newfloat}

\usepackage{amssymb}
\usepackage{bm}
\usepackage{booktabs}
\usepackage{multirow}
\usepackage{rotating}
\usepackage{amsmath}

\usepackage{listings}
\DeclareCaptionStyle{ruled}{labelfont=normalfont,labelsep=colon,strut=off} 
\floatstyle{ruled}
\newfloat{listing}{tb}{lst}{}
\floatname{listing}{Listing}
\title{ParaGuide: Guided Diffusion Paraphrasers for \\Plug-and-Play Textual Style Transfer}
\author {
    Zachary Horvitz\textsuperscript{\rm 1},
    Ajay Patel\textsuperscript{\rm 2},
    Chris Callison-Burch\textsuperscript{\rm 2},
    Zhou Yu\textsuperscript{\rm 1},
    Kathleen McKeown\textsuperscript{\rm 1}
}
\affiliations {
    \textsuperscript{\rm 1} Columbia University \\
    \textsuperscript{\rm 2} University of Pennsylvania \\
    zfh2000@columbia.edu, 
    ajayp@seas.upenn.edu,
    ccb@seas.upenn.edu,
    zy2461@columbia.edu,
    kathy@cs.columbia.edu

}

\usepackage{bibentry}

\begin{document}

\maketitle

\begin{abstract}
 Textual style transfer is the task of transforming stylistic properties of text while preserving meaning. Target ``styles" can be defined in numerous ways, ranging from single attributes (e.g. formality) to authorship (e.g. Shakespeare). 
    Previous unsupervised style-transfer approaches generally rely on significant amounts of labeled data for only a fixed set of styles or require large language models. In contrast, we introduce a novel diffusion-based framework for general-purpose style transfer that can be flexibly adapted to arbitrary target styles at inference time. Our parameter-efficient approach, \textsc{ParaGuide}, leverages paraphrase-conditioned diffusion models alongside gradient-based guidance from both off-the-shelf classifiers and strong existing style embedders to transform the style of text while preserving semantic information. We validate the method on the Enron Email Corpus, with both human and automatic evaluations, and find that it outperforms strong baselines on formality, sentiment, and even authorship style transfer.
\end{abstract}

\section{Introduction}

Diffusion models \cite{sohldickstein2015deep,ho2020denoising,song2022denoising} were originally popularized for image synthesis \cite{pmlr-v139-nichol21a, saharia2022palette}. More recently, however, diffusion has been successfully applied to text. Diffusion-based language models are increasingly competitive with traditional approaches for text generation \cite{li2022diffusionlm, gulrajani2023likelihoodbased, han2023ssdlm, han2023ssd2}, and on text-to-text modeling tasks \cite{mahabadi2023tess, yuan2023seqdiffuseq}. 

A key benefit of diffusion language models for text is their high degree of controllability. Diffusion-based approaches hierarchically denoise a continuous representation of an entire sequence, and this process can be effectively guided with gradient-based methods \cite{li2022diffusionlm,han2023ssdlm,gulrajani2023likelihoodbased}.  
This differs from the dominant approach of \textit{autoregressive} decoding, where text is generated by sequentially sampling tokens. Steering pretrained autoregressive models has proven difficult, as their text is  greedily decoded and guidance must operate on partial sequences \cite{li2022diffusionlm, dathathri2020plug, krause2020gedi, Yang_2021}. 
\begin{figure}[t]
\includegraphics[width=8cm]{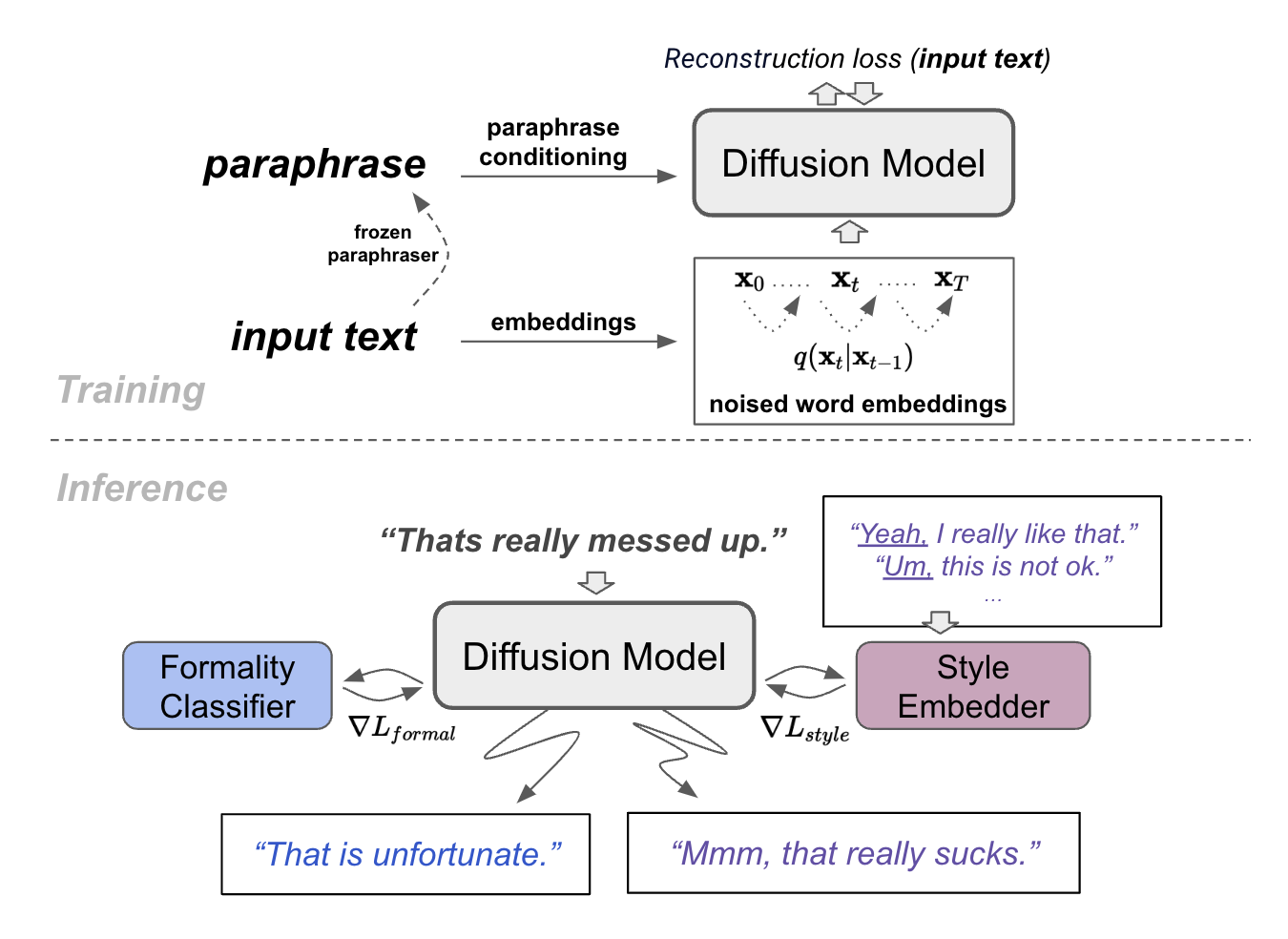}
\centering
\caption{We train paraphrase-conditioned text diffusion models to reconstruct semantically consistent text from noised word embeddings. At inference time, we guide the reconstruction towards target styles with off-the-shelf models.}
\label{fig:figure1}
\end{figure}
We leverage the controllability of nascent text diffusion methods and adapt them to \textit{style transfer}.

In textual style transfer, the objective is to transform the style of the text to exhibit an attribute (such as ``formality"), or a target author's style, while preserving meaning \cite{jin-etal-2022-deep, krishna2020reformulating, patel2022lowresource}. 
The scarcity of style-transfer datasets has motivated \textit{unsupervised} style transfer approaches that perform attribute and authorship style transfer without paired data. 
These approaches generally require retraining for new target styles. 

In contrast, we introduce a plug-and-play diffusion framework for unsupervised style transfer.\footnote{Our code is publicly available at \url{https://github.com/zacharyhorvitz/ParaGuide}.} We initially train a text diffusion model to reconstruct semantically consistent text from paraphrases, but at inference time, we perform new attribute or authorship style transfers by guiding reconstruction with gradients from off-the-shelf models (Figure \ref{fig:figure1}). This allows users to leverage the numerous text classifiers on platforms like Hugging Face\footnote{\url{https://huggingface.co/models}} to specify target styles.
 Beyond guidance from classifiers, our method enables bringing recent advances in representation learning to bear by ``plugging in" authorship representations like Style Embeddings \cite{wegmann-etal-2022-author} and Universal Authorship Representations \cite{rivera-soto-etal-2021-learning}. This enables our approach to perform challenging tasks like \textit{low-resource authorship style transfer} \cite{patel2022lowresource}.
 
 %

\textbf{Our contributions are as follows:}

    \begin{enumerate}
        \item We propose a novel framework for textual style transfer based on paraphrase-conditioned diffusion models, \textsc{ParaGuide}.

        \begin{itemize}
            \item Unlike existing style-transfer approaches, this framework enables gradient-based guidance using off-the-shelf models at inference time. 
            \item Beyond classifier guidance, we show that existing authorship representations can be plugged in for control. Even with limited available data, this allows \textsc{ParaGuide} to competitively perform authorship style transfer.
            \item Style transfer requires balancing style-transfer accuracy with fluency and meaning preservation. Our framework enables explicit control over this trade-off through varying guidance strength ($\lambda$).
        \end{itemize}
        \item We validate our approach on formality and sentiment transfer, where it outperforms strong baselines on automatic evaluations. Additionally, we perform a human evaluation for formality transfer. 
        \item \textsc{Paraguide} represents early work exploring the promising benefits afforded by  text diffusion models. To our knowledge, we are the first to adapt these approaches to unsupervised textual style transfer.

    \end{enumerate}

\section{Related Work}

Other unsupervised transfer approaches, like \textsc{Strap}, create pseudo-parallel corpora by corrupting texts to remove stylistic attributes, then training models to reconstruct the uncorrupted text \cite{krishna2020reformulating, riley-etal-2021-textsettr, ma-etal-2020-powertransformer}. These approaches cannot use new stylistic representations without retraining and do not incorporate control from off-the-shelf models.
Additionally, \textsc{Strap} has been shown to require large amounts of style-specific training data \cite{patel2022lowresource}.
 Prior work has explored applying controllable text generation techniques to style transfer \cite{dale2021text, kumar2021controlled, mireshghallah2022mix}. Our approach is most similar in spirit to  \citet{mireshghallah2022mix}. Their approach is also learning free and non-autoregressive, but performs a discrete search which is very computationally expensive for long sequences, cannot leverage the rich information in gradients, and confines the search space at each step to token-level substitutions. 
Recently, the emergent ability of Large Language Models (LLMs) to perform in-context learning \cite{brown2020language} has presented formidable baselines for text generation and style transfer \cite{reif-etal-2022-recipe, patel2022lowresource}. Unlike LLMs, \textsc{ParaGuide} allows gradient-based control and can leverage stylistic embeddings, and is not restricted to brittle guidance through text-based prompts. Moreover, these approaches typically require models with billions of parameters \cite{Radford2019LanguageMA}.

\section{ParaGuide}

\subsection{Overview}

\textsc{ParaGuide} has three primary steps:

\begin{enumerate}
    \item Generating an \textbf{initial paraphrase} of an input text with an autoregressive (AR) model.
    \item Using a paraphrase-conditioned text \textbf{diffusion model} to iteratively reconstruct the input text from this paraphrase over a number of diffusion steps.    
    \item At each diffusion step, computing gradients for arbitrary differentiable losses, and using these gradients for \textbf{guidance} towards a target style.

\end{enumerate}

Here, we first use paraphrasing to generate an intermediate text that is semantically consistent with the input text but without the original stylistic attributes \cite{krishna2020reformulating}. We then reconstruct the text with a paraphrase-conditioned diffusion model. During reconstruction, we optimize some loss function specified by a guidance model \cite{han2023ssdlm, li2022diffusionlm, gulrajani2023likelihoodbased}. The result is a semantically consistent output in the desired target style.

\subsection{Initial Paraphrase Generation}

At both training and inference time, \textsc{ParaGuide} requires \textit{(paraphrase, original text)} pairs. 
To generate this synthetic data, we leverage an existing, publicly available model \cite{zhang2020pegasus},  specifically fine-tuned for paraphase generation. We include additional information describing this procedure in our Appendix. This aspect of our approach distills performant, but less controllable, autoregressive paraphrasers into controllable diffusion models.



\subsection{Paraphrase-Conditioned Diffusion}

In this section, we introduce the components of our paraphrase-conditioned text diffusion model.

\subsubsection{Diffusion}

Diffusion approaches \cite{sohldickstein2015deep, ho2020denoising, song2022denoising}, consist of two Markov chains, a \textit{forward process} and a \textit{reverse process}. In the \textit{forward process}, the original data $\mathbf{x}_0$ is converted to pure Gaussian noise by incrementally adding noise over multiple discrete time steps, $\{0,...,T\}$. Each of these intermediate noised latents, $\mathbf{x}_t$, can be directly sampled as follows: 
\begin{equation}
    \mathbf{x}_t = \sqrt{\bar{a}_t} \mathbf{x}_0 + \sqrt{1-\bar{a}_t} \bm{\epsilon}_t; \bm{\epsilon}_t\sim \mathcal{N}(0, \mathbf{I}),
\end{equation}
where $\bm{\epsilon}_t$ is random noise and $\bar{a}_t$ specifies a well-behaved schedule such that $\bar{a}_t \rightarrow 0$ as $t \rightarrow T$.
The \textit{reverse process} is parameterized by a model, which is trained to reconstruct the original data from pure noise ($\mathbf{x}_T$) by iteratively estimating $\bm{\epsilon}_t$ (or equivalently $\mathbf{x}_0$) and working backwards in time, from $t=T$ to $t=0$. 

In the image domain, pixels are used as the representation of $\mathbf{x}_0$. In contrast, text is discrete, and the underlying continuous domain is less obvious. Several existing text diffusion approaches operate on word embeddings \cite{li2022diffusionlm, yuan2023seqdiffuseq, gulrajani2023likelihoodbased}, while others noise token logit simplexes, like SSD-LM \cite{han2023ssdlm, han2023ssd2, mahabadi2023tess}. \textsc{ParaGuide} performs diffusion in word embedding space, but incorporates several benefits of simplex methods. 

\subsubsection{Categorical Reparameterization}

While diffusion with word logits has several desirable properties \cite{han2023ssdlm}, logits are a high dimension latent representation (sequence length $\times$ vocabulary size), which makes both training and inference slower and more memory intensive than operating directly on the word embedding space.
Also, unlike the probability simplex, pretrained word embedding spaces are well-suited for meaning-preserving style transfer, as neighbors are often semantically similar \cite{mikolov2013efficient}.\footnote{Additionally, we observed that the original SSD-LM approach of adding Gaussian noise to logits before a $\operatorname{softmax}$ operation results in almost completely noised text early on in the forward process, and early steps could be discarded with little effect on our final outputs.} 
As a result, in \textsc{ParaGuide}, we employ noised word embeddings for our latent representations, and define our forward process as:

\begin{equation}
\mathbf{x}_t = \sqrt{\bar{a}_t} E(\mathbf{w}) + \sqrt{(1-\bar{a}_t)} \bm{\epsilon}_t, 
\end{equation}
\noindent
where $\mathbf{w}$ is our original text and $E$ is an embedding lookup. Rather than directly estimate the original word embeddings in our reverse process, however, we estimate $E(\mathbf{w})$ with a diffusion model that first outputs a posterior over discrete tokens, like \citet{gulrajani2023likelihoodbased}:

\begin{equation}
\hat{\mathbf{w}}_t \sim p_{\theta}(. | \mathbf{x}_t, t, \mathbf{p}),
\end{equation}
where $\mathbf{x}_t$ is our noised embedding, and $\mathbf{p}$ is our input paraphrase. We sample intermediate tokens from this distribution like in SSD-LM \cite{han2023ssdlm}, and these tokens are embedded for the next $t_{n-1}$th diffusion step:
\begin{equation}
\mathbf{x}_{t-1} = \sqrt{\bar{a}_{t-1}} E(\hat{\mathbf{w}}_t) + \sqrt{(1-\bar{a}_{t-1})} \bm{\epsilon};  \bm{\epsilon} \sim \mathcal{N}(0, \mathbf{I}) 
\end{equation}
\noindent This approach still provides the controllability benefits of SSD-LM, as gradient-based control can be applied to the intermediate token predictions, which we will discuss in the Guidance section.

\subsubsection{Diffusion Model Architecture}

We build on the SSD-LM architecture \cite{han2023ssdlm}, which uses a bidirectional RoBERTa encoder \cite{liu2019roberta} to output token probabilities at each diffusion step, conditioned on a noised representation and timestep.  However, we make several changes to adapt their simplex-diffusion approach for text-to-text tasks like paraphrasing.  First, as noted in the previous section, we modify their model to operate on noised word embeddings, rather than word logits. Additionally, as in \citet{mahabadi2023tess}, we also modify the original semi-autoregressive approach to be entirely diffusion-based. Finally, like other text-to-text diffusion approaches \cite{mahabadi2023tess, yuan2023seqdiffuseq}, we condition on an input (in our case, the paraphrase, $\mathbf{p}$), by concatenating it with our noised latent representation.
Unlike these approaches, we incorporate stylistic guidance, as outlined in the Guidance section.

\subsubsection{Diffusion Model Loss}

Following \citet{han2023ssdlm} and \citet{mahabadi2023tess}, we train the diffusion model by minimizing the cross entropy between the model's posterior at each diffusion timestep and the ground-truth tokens $\mathbf{w}$, but given the timestep $t$, noised embeddings $\mathbf{x}_t$, and paraphrase $\mathbf{p}$:

\begin{equation}
 \mathcal{L}(\theta) = \mathbb{E}_{t \sim \mathcal{U}(1,T)}[-\log{p_\theta}(\mathbf{w} | \mathbf{x}_t, t, \mathbf{p} )]  
\end{equation}

\subsubsection{Diffusion Noise Schedule}

Several approaches to diffusion language modeling \cite{han2023ssdlm, mahabadi2023tess, han2023ssd2} have repurposed the \textit{cosine} schedule \cite{pmlr-v139-nichol21a} from computer vision, while others have adopted the \textit{sqrt} schedule \cite{li2022diffusionlm, yuan2023seqdiffuseq}. In contrast, we train \textsc{Paraguide} with a dramatically less aggressive noise schedule:
\begin{equation}
    \bar{a}_t = \sqrt{\frac{T-t}{T}}
\end{equation}

\noindent This schedule falls to zero much more slowly than the cosine and sqrt schedules, destroying information less quickly. The schedule is motivated by our observation that skipping early steps with the cosine schedule had no noticeable effect on model outputs, and experiments that showed improved fluency and meaning preservation.\footnote{We view less aggressive noise schedules as a possible alternative to self-conditioning \cite{strudel2022selfconditioned, han2023ssd2, mahabadi2023tess}, which handles information loss by also conditioning diffusion language models on previous $x_{t-1}$ predictions. We found that self-conditioning improved fluency but hurt control.}

\subsubsection{Diffusion Model Inference}

At inference time, we first generate a paraphrase, $\mathbf{p}$ of our input text. We then
sample initial noise $\mathbf{x}_T \sim \mathcal{N}(0,I)$. For each step in the reverse process ($t \in [T,1]$), we then compute token logits using our model:
\begin{equation}
\mathbf{l}_{t} = \operatorname{logits}_{\theta}(. | \mathbf{x}_t, t, \mathbf{p})
\end{equation}

We then sample from the model's posterior:\footnote{This is similar to the sampling approach used by \citet{han2023ssdlm} and clamping in \citet{li2022diffusionlm}}

\begin{equation}
\hat{\mathbf{w}}_{t} \sim \operatorname{top-p}(\operatorname{softmax}(\bf{l}_{t}))
\end{equation}

\noindent After sampling $\hat{\mathbf{w}}_{t}$, we iteratively work backwards in time by embedding these tokens using the word embedding lookup, $E$, and then adding noise to produce $\mathbf{x}_{t-1}$, the latent for the previous diffusion timestep, following \citet{han2023ssdlm}:

\begin{equation}
\mathbf{x}_{t-1} = \sqrt{\bar{a}_{t-1}} E(\hat{\mathbf{w}}_{t}) + \sqrt{(1-\bar{a}_{t-1})} \bm{\epsilon}; \bm{\epsilon} \sim \mathcal{N}(0,I)
\end{equation}

In this fashion, the model starts from random noise and an input paraphrase, and then iteratively generates a semantically consistent output text. A critical advantage of applying diffusion models to this task is that we can use gradient-based guidance to steer our outputs towards specific target styles. We discuss this in the next section.

\subsection{Guidance}

\textsc{Paraguide} can incorporate guidance from any model that is 1) differentiable and 2) uses the same tokenization scheme as the base diffusion paraphraser:

\begin{equation}
  \mathbf{l}_{t} = \mathbf{l}_{t,\textit{init}} - \lambda \nabla_{\mathbf{l}_t} L_{\textit{guidance}}(\mathbf{l}_{t,\textit{init}})
\end{equation}
where $\mathbf{l}_{t,\textit{init}}$ are the initial logit predictions at timestep $t$, and $L_{\textit{guidance}}$ specifies a guidance loss.

Because our diffusion model employs a RoBERTa \cite{liu2019roberta} tokenization scheme, we can incorporate guidance from the many available models built on the popular RoBERTa encoder backbone. We explore two forms of guidance loss for style transfer: The first is based on attribute classifiers, and the second is based on distances in stylistic embedding space.

\subsubsection{Attribute Classifiers}

Following  \citet{han2023ssdlm}, we use a classifier, $f_\phi(\cdot)$ to generate texts with a target attribute, $y$, by applying drift to the full sequence of logits, $\mathbf{l}_t$, at each intermediate diffusion step:

\begin{equation}
    L_{\textit{guidance}}(\mathbf{l}_t) = -log(f_\phi(y | \mathbf{l}_t))
\end{equation}

Additionally, like \citet{han2023ssdlm}, we can trivially adapt classifiers to accept logits, rather than word embeddings, by using the $\operatorname{softmax}$ function with some temperature, $\tau$, to compute a probability simplex over the vocabulary. We can then project with the classifier's embedding lookup, $E_{\phi}$:

\begin{equation}
 \mathbf{\tilde{e}}_{\phi,t} = \operatorname{softmax}(\frac{\mathbf{l}_t}{\tau}) \times E_{\phi}
\end{equation}

\begin{algorithm}[tb]
\caption{ParaGuide Style Transfer}
\label{alg:algorithm}
\textbf{Input}: \\ Input Text in Source Style $\mathbf{w}$,  \\ Guidance Loss $L_{\textit{guidance}}$,  Guidance Strength $\lambda$ \\
\textbf{Output}: Output Text in the Target Style \\
\begin{algorithmic}[1] 
\STATE $\mathbf{p} = \textit{paraphraser}(\mathbf{w})$
\STATE $\mathbf{x}_T \sim \mathcal{N}(0,1)$
\FOR{$t = T,...,1$}
\STATE $\mathbf{l}_{t,\textit{init}} = \operatorname{logits}_{\theta}(. | \mathbf{x}_t, t, \mathbf{p})$
\IF {$ \lambda \neq 0 $}
\FOR{$i = 1,...,k$}
\STATE $\mathbf{l}_{t} \leftarrow \mathbf{l}_{t} -  \lambda \cdot \sin(\pi\frac{t}{T}) \cdot \nabla_{\mathbf{l}_{t,\textit{init}}}L_{\textit{guidance}}(\mathbf{l}_{t,\textit{init}})$
\ENDFOR
\ENDIF
\STATE $\hat{\mathbf{w}}_{t} \sim \operatorname{top-p}(\operatorname{softmax}(\bf{l}_t))$
\STATE $\bm{\epsilon} \sim \mathcal{N}(0,I)$
\STATE $\mathbf{x}_{t-1} = \sqrt{\bar{a}_t} E(\hat{\mathbf{w}}_{t}) + \sqrt{(1-\bar{a}_t)} \bm{\epsilon}$
\ENDFOR
\STATE \textbf{return} $\mathbf{w}_0$
\end{algorithmic}
\end{algorithm}

This results in a linear combination of word embeddings at each timestep $\mathbf{\tilde{e}}_{\phi,t}$, based on each token's assigned probability mass. These embeddings are passed to the attribute model, and gradients are computed through them to increase or decrease the probabilities of different tokens to maximize the probability of attribute $y$. In contrast to SSD-LM \cite{han2023ssdlm}, \textsc{ParaGuide} applies this drift to a diffusion model trained to reconstruct semantically consistent text, which enables meaning-preserving style transfer. We can balance the trade-off between semantic consistency and style transfer by varying $\lambda$.

\subsubsection{Style Embedding Distance}

For authorship style transfer, we take the novel approach of leveraging guidance from stylistic embedding models, including Style Embeddings \cite{wegmann-etal-2022-author}, that are constrastively trained to identify authorship styles. 



To guide our paraphrases with a style embedder, $g_\phi$, we compute the gradient of $\mathbf{l}_{t}$ with respect to its average distance in style embedding space to the target author's $n$ texts, $[\mathbf{y_1},\mathbf{y_2}...,\mathbf{y_n}]$:

\begin{equation}
L_{\textit{guidance}}(\mathbf{l}_t) =\frac{\sum_{i=1}^{n} d(g_{\phi}(\mathbf{l}_{t}),g_{\phi}(\mathbf{y_i}))}{n}
\end{equation}

We use cosine distance for $d(\cdot)$. At every diffusion step, by minimizing the distance in style embedding space, we steer the output text towards the target author's style.

\subsubsection{Guidance Schedule}

We observed that using the same $\lambda$ for control at all diffusion timesteps leads to disfluent solutions. Intuitively, large gradient steps at the end of the reverse diffusion process are undesirable, as optimizing the control objective can lead to ungrammatical text. Simultaneously, large steps early on in the reverse process are also undesirable, as these initial predictions are generally incoherent, and out of distribution for off-the-shelf models.

As a result, for all forms of guidance, we employ a sinusoidal schedule for controlling drift:

\begin{equation}
\lambda_t = \lambda \cdot \sin(\pi\frac{t}{T})
\end{equation}

This increases and then anneals the strength of drift during the reverse process. Additionally, we make $k$ gradient updates per diffusion step, like \citet{li2022diffusionlm}. \textsc{ParaGuide}'s complete inference procedure is specified in Algorithm \ref{alg:algorithm}.

\section{Experimental Setup}

\subsection{Dataset}

We evaluate our method on the Enron Email Corpus, which comprises several hundred thousand emails made public during the US government's investigation of Enron \cite{Klimt2004TheEC,peterson-etal-2011-email}. The dataset contains emails from the inboxes of 150 Enron employees, sent from over one thousand accounts.

The Enron corpus presents an ideal testbed for plug-and-play style transfer of both authorship and attributes. For the former, email meta-data enables attributing messages to specific authors for authorship transfer.  The emails also present diverse stylistic attributes, including different degrees of formality \cite{peterson-etal-2011-email} and divergent rhetorical styles \cite{stylistic-variation}. 

Ultimately, we need to evaluate whether our email style transfer approach generalizes to new authors and texts. Therefore, we randomly select $10\%$ of addresses to be the holdout authors for both authorship and attribute evaluations. These $110$ authors present a low-resource authorship corpus, as the median holdout author has only $23$ emails. For our authorship experiments, we evaluate each approach by selecting up to $5$ test emails per holdout source author, and transferring these to $5$ other random holdout authors. 

To build our training and validation datasets for attribute style transfer, we use popular existing  formality and sentiment classifiers to score texts from the holdout authors in the Enron dataset. Critically, we set aside these external classifiers and avoid using them as guidance for \textsc{ParaGuide} at inference time.
In addition to the Enron corpus, we also build a pretraining corpus from the Reddit Million User Dataset (MUD) \cite{andrews2019learning, khan2021deep}, which includes $4$ million comments by $400$k different Reddit users. We use the same paraphrasing procedure on both the Enron and Reddit datasets to generate (\textit{paraphrase}, \textit{original text}) training pairs.

\subsection{Implementation Details}

To train our diffusion model, we fine-tuned the publicly available SSD-LM RoBERTa-Large checkpoint\footnote{\url{https://huggingface.co/xhan77/ssdlm}} \cite{han2023ssdlm} with our previously stated modifications to the architecture and noise schedule. We first fine-tune the diffusion model on Reddit paraphase pairs, and then continue fine-tuning on the Enron non-holdout author paraphrase pairs. We fine-tune all parameters except the word embedding lookup. Additional implementation details are included in our Appendix. 

\subsection{Baselines}

\subsubsection{Attribute Style Transfer}

For attribute transfer, we compare to Mix and Match (M\&M) \cite{mireshghallah2022mix} and consider both the \textsc{Disc} and \textsc{Ham} configurations from the original paper \cite{mireshghallah2022mix}. To better compare with our approach, however, we replace the original BERT model with RoBERTa-large and also include results where we fine-tune this model on Enron Email training data. 

We also implement a \textsc{Strap} baseline \cite{krishna2020reformulating} with pretrained T$5$-Large models \cite{raffel2020exploring}, fine-tuned on Reddit and then Enron paraphrase pairs. In contrast to M\&M and \textsc{ParaGuide}, which are learning-free approaches, \textsc{Strap} requires training attribute-specific models on the Enron data classified by the external classifiers. We fine-tune four STRAP models for informality, formality, positive sentiment, and negative sentiment.

\subsubsection{Authorship Style Transfer}

For the task of authorship style transfer on the Enron Email Corpus, we consider \textsc{Strap} \cite{krishna2020reformulating}, and the \textsc{Bert}, \textsc{Ling}, and \textsc{Para} approaches from \citet{patel2022lowresource}. We also consider a ChatGPT-3.5 style transfer approach, where we prompt the model with up to $16$ in-context examples of a target author's style.
In contrast to our other approaches, we fine-tune $110$ author-specific \textsc{Strap} models on $60\%$ of each holdout author's data.

\subsection{Evaluation Metrics}

\subsubsection{Attribute Style Transfer}

Following \citet{, mireshghallah2022mix}, we measure style transfer accuracy with two classifiers. First, 
\textit{Internal Accuracy} measures the style transfer accuracy of the classifier used at inference time by Mix and Match and \textsc{ParaGuide}.
In contrast, \textit{External Accuracy} measures the style transfer accuracy using a classifier set aside for evaluation. 

We measure textual \textit{Similarity} by  computing Mutual Implication Score (MIS) \cite{babakov-etal-2022-large} and \textit{Fluency} with a model trained on the CoLA dataset \cite{morris2020textattack,warstadt-etal-2019-neural}.\footnote{textattack/roberta-base-CoLA} For an aggregate metric of model performance, we  compute a \textit{Joint} metric by taking the sentence-wise geometric mean of \textit{External Accuracy}, \textit{Similarity}, and \textit{Fluency}, similar to \citet{krishna2020reformulating}. 

For formality transfer, we additionally run a human evaluation of style-transfer approaches that scored highest on our automatic evaluations. We asked annotators to compare model outputs to the reference inputs, and score ($\{0,1\}$) their \textit{Similarity}, \textit{Fluency}, and \textit{Formality}. We include additional details describing our human evaluations in the Appendix.

\subsubsection{Authorship Style Transfer}

To evaluate authorship style transfer, we adopt the \textit{Confusion} metric from the evaluation framework defined by \citet{patel2022lowresource}, where the authors utilize pretrained style embedders \cite{wegmann-etal-2022-author,rivera-soto-etal-2021-learning} to measure style transfer success. \textit{Confusion}, which is similar to style transfer accuracy, is the percentage of the time that the style transfer output is closer to the target author than the source author in representational embedding space. As with attribute transfer, we similarly compute \textit{Similarity} and \textit{Fluency}, and \textit{Joint}, but use \textit{Confusion} in place of transfer accuracy.

We compute the above metrics for both Style Embeddings \cite{wegmann-etal-2022-author} and Universal Authorship Representations (UAR) \cite{rivera-soto-etal-2021-learning}. Similar to our external style classifier for attribute transfer, UAR provides a holdout embedding space that \textsc{Paraguide} does not directly optimize at inference time.

\vspace{18pt}

\begin{table*}[t]
\fontsize{9}{11}\selectfont
\centering

\begin{tabular}{lccccc}
\toprule
Method & Int. Acc ($\rightarrow \textit{F}$,$\rightarrow \textit{I}$) & Ext. Acc ($\rightarrow \textit{F}$,$\rightarrow \textit{I}$) &  Sim ($\rightarrow \textit{F}$,$\rightarrow \textit{I}$) & Fluency ($\rightarrow \textit{F}$,$\rightarrow \textit{I}$) & Joint ($\rightarrow \textit{F}$,$\rightarrow \textit{I}$)  \\
\midrule
STRAP$_{\textit{fine-tuned}}$ & 0.45 (0.8, 0.1) & 0.45 (0.76, 0.13) &  0.50 (0.54, 0.47) & \textbf{0.73} (0.75, 0.71) & 0.31 (0.54, 0.08) \\
\midrule
M\&M (Disc) &  0.63 (0.59, 0.67) & 0.55 (0.44, 0.65) &  0.24 (0.19, 0.3) & 0.62 (0.62, 0.62) & 0.23 (0.19, 0.27) \\
M\&M (Hamming)  & 0.58 (0.59, 0.57) & 0.51 (0.46, 0.57) &  0.40 (0.29, 0.52) &  0.61 (0.61, 0.6) & 0.26 (0.21, 0.31) \\
M\&M$_{enron}$ (Disc)  & 0.58 (0.62, 0.55) & 0.51 (0.47, 0.56) & 0.31 (0.26, 0.37) & 0.61 (0.61, 0.61) & 0.24 (0.22, 0.26) \\
M\&M$_{enron}$ (Hamming)&  0.51 (0.56, 0.46) & 0.47 (0.45, 0.48) & 0.45 (0.35, 0.55) & 0.62 (0.62, 0.62) & 0.25 (0.23, 0.28) \\
\midrule

PGuide ($\lambda=1 \mathrm{e}4$) & \textbf{0.97} (0.96, 0.99) & \textbf{0.83} (0.68, 0.99) &  0.40 (0.37, 0.44) & 0.55 (0.59, 0.51) & 0.45 (0.37, 0.53) \\
PGuide ($\lambda=5 \mathrm{e}3$) & 0.97 (0.96, 0.98) & 0.82 (0.65, 0.99) &  0.40 (0.36, 0.45) & 0.56 (0.59, 0.52) & 0.45 (0.37, 0.53) \\
PGuide ($\lambda=1 \mathrm{e}3$) & 0.95 (0.93, 0.98) & 0.81 (0.64, 0.98) &  0.45 (0.4, 0.49) &  0.60 (0.62, 0.57) & 0.47 (0.37, 0.56) \\
PGuide ($\lambda=5 \mathrm{e}2$) & 0.94 (0.9, 0.97) & 0.81 (0.63, 0.98) &  0.47 (0.44, 0.5) & 0.61 (0.64, 0.58) & 0.48 (0.39, 0.58) \\
PGuide ($\lambda=2 \mathrm{e}2$) & 0.91 (0.85, 0.98) & 0.76 (0.58, 0.95) &  \textbf{0.52} (0.5, 0.53) & 0.63 (0.65, 0.61) & \textbf{0.48} (0.38, 0.59) \\
\bottomrule
\end{tabular}
\caption{Automatic Formality Evaluations. We report accuracy for both the \textit{Internal} and  \textit{External} classifiers. The best results are bolded. We also decompose results into formality ($\rightarrow F$) and informality ($\rightarrow I$) transfer.  }
\label{table:formal_auto}
\end{table*}
\begin{table*}[t]
\fontsize{9}{11}\selectfont
\centering
\begin{tabular}{lcccccc}
\toprule
Method & Accuracy ($\rightarrow \textit{F}$,$\rightarrow \textit{I}$) &  Sim ($\rightarrow \textit{F}$,$\rightarrow \textit{I}$) & Fluency ($\rightarrow \textit{F}$,$\rightarrow \textit{I}$) & Joint ($\rightarrow \textit{F}$,$\rightarrow \textit{I}$) & \\
\midrule
STRAP$_{\textit{fine-tuned}}$  & 
0.51 (0.10,  0.91) & 0.35 (0.32, 0.37) & 0.03 (0.04, 0.01) & 0.00 (0.00, 0.00)

\\
M\&M (Hamming) & 0.47 (0.14, 0.80) & 0.49 (0.31, 0.67) & 0.46 (0.27, 0.64) & 0.20 (0.03, 0.36) \\
PGuide ($\lambda=2 \mathrm{e}2$) & \textbf{0.65} (0.39, 0.90) & \textbf{0.58} (0.54, 0.61) & \textbf{0.69} (0.61, 0.77) & \textbf{0.33} (0.23, 0.43)\\
\bottomrule
\end{tabular}
\caption{Human Formality Evaluations. We asked annotators to rate outputs from models with the highest automatic scores as formal or informal (\textit{Accuracy}), whether their meaning was similar to the original (\textit{Similarity}), and whether the outputs were well-formed/grammatical (\textit{Fluency}). \textit{Joint} aggregates these scores together at the sentence-level. }
\label{table:formal_human}
\end{table*}

\section{Results}

\subsection{Attribute Style Transfer}

In this section, we review our evaluation results for attribute transfer. We include representative outputs in the Appendix.  

\subsubsection{Automatic Evaluations}

\begin{table*}[t]
\fontsize{9}{11}\selectfont
\centering
\begin{tabular}{lcccccc}

\toprule
Method & Int. Acc ($\rightarrow \textit{P}$,$\rightarrow \textit{N}$) & Ext. Acc ($\rightarrow \textit{P}$,$\rightarrow \textit{N}$) &  Sim ($\rightarrow \textit{P}$,$\rightarrow \textit{N}$) & Fluency ($\rightarrow \textit{P}$,$\rightarrow \textit{N}$) & Joint ($\rightarrow \textit{P}$,$\rightarrow \textit{N}$) & \\
\midrule
STRAP$_{\textit{fine-tuned}}$ & 0.11 (0.16, 0.05) & 0.29 (0.38, 0.19) &   0.5 (0.5, 0.49) & \textbf{0.74} (0.72, 0.76) & 0.18 (0.24, 0.12) \\
\midrule
M\&M (Disc) &  0.2 (0.01, 0.38) &  0.5 (0.32, 0.67) & 0.34 (0.46, 0.22) & 0.63 (0.62, 0.64) & 0.21 (0.17, 0.25) \\
M\&M (Ham) & 0.14 (0.02, 0.26) & 0.39 (0.23, 0.55) & 0.45 (0.58, 0.32) &  0.62 (0.6, 0.63) & 0.19 (0.14, 0.24) \\
M\&M$_{enron}$ (Disc)  & 0.1 (0.02, 0.18) & 0.38 (0.29, 0.47) &  0.4 (0.48, 0.33) &  0.62 (0.6, 0.64) & 0.19 (0.16, 0.22) \\
M\&M$_{enron}$ (Ham) &  0.08 (0.02, 0.13) & 0.31 (0.21, 0.41) &  \textbf{0.52} (0.6, 0.44) & 0.62 (0.61, 0.64) &  0.16 (0.13, 0.2) \\
\midrule

PGuide ($\lambda=1 \mathrm{e}4$) & \textbf{0.73} (0.78, 0.68) &  \textbf{0.8} (0.86, 0.74) &  0.13 (0.2, 0.06) & 0.43 (0.43, 0.43) &  0.2 (0.27, 0.13) \\
PGuide ($\lambda=5 \mathrm{e}3$) & 0.7 (0.76, 0.65) & 0.79 (0.87, 0.71) & 0.15 (0.22, 0.07) & 0.43 (0.45, 0.41) &  0.22 (0.3, 0.14) \\
PGuide ($\lambda=1 \mathrm{e}3$) & 0.65 (0.75, 0.54) & 0.75 (0.81, 0.69) & 0.25 (0.32, 0.18) & 0.48 (0.53, 0.43) & 0.28 (0.35, 0.21) \\
PGuide ($\lambda=5 \mathrm{e}2$) & 0.57 (0.71, 0.43) & 0.68 (0.74, 0.62) & 0.33 (0.37, 0.28) & 0.51 (0.55, 0.47) & 0.29 (0.35, 0.23) \\
PGuide ($\lambda=2 \mathrm{e}2$) & 0.35 (0.47, 0.22) & 0.56 (0.61, 0.51) &  0.42 (0.44, 0.4) & 0.59 (0.64, 0.55) & \textbf{0.29} (0.33, 0.25) \\
\bottomrule
\end{tabular}

\caption{Automatic Sentiment Evaluations. Like for the formality results, we break down scores into positive ($\rightarrow P$) and negative ($\rightarrow N$) transfer, and report scores for both the \textit{Internal} and  \textit{External} classifiers. }
\label{table:sentiment_auto}
\end{table*}
Tables \ref{table:formal_auto} and \ref{table:sentiment_auto} present our automatic evaluation results for formality and sentiment transfer. For each approach, we display the average score for each metric, along with the breakdown for formal/informal $(\rightarrow F, \rightarrow I)$ and positive/negative $(\rightarrow P, \rightarrow N)$. 

\textsc{ParaGuide} outperforms all other approaches on all aggregate \textit{Joint} metrics, across both sentiment and formality experiments. Additionally, \textsc{ParaGuide} significantly surpasses all baselines on transfer accuracy. 
Despite the inherent trade-off between transfer accuracy and meaning preservation, on formality, \textsc{ParaGuide} ($\lambda=2\mathrm{e}2$) outperforms all baseline approaches on both transfer accuracy \textit{and} meaning preservation. On sentiment transfer, \textsc{ParaGuide}'s increased accuracy incurs a larger cost to semantic similarity, but this is expected in successful sentiment transfer, which involves changing the polarity of texts \cite{jin-etal-2022-deep}.

\subsubsection{Human Evaluation}

Table \ref{table:formal_human} displays the results of our human formality evaluation, where annotators rated the \textit{Formality}, \textit{Similarity}, and \textit{Fluency} of model outputs. When evaluated by humans, \textsc{ParaGuide} significantly outperforms the top performing baselines across all aggregate metrics ($p=0.05$). Notably, this is even true for the \textit{Fluency} metric, where annotators rated whether outputs were reasonable, coherent emails. This result was unexpected given \textsc{ParaGuide}'s comparatively unimpressive automatic \textit{Fluency} scores, but could be explained by differences between email writing practices and the composition of the CoLA training corpus \cite{warstadt-etal-2019-neural}. In contrast, the \textsc{Strap} baseline dramatically underperforms on our human evaluation. Manually inspecting outputs, we found that the \textsc{Strap} models we fine-tuned for attribute transfer generate highly repetitive text. We suspect that this results from fine-tuning on our limited dataset, and aligns with previous work, which has shown that \textsc{Strap}'s performance is heavily reliant on dataset size \cite{patel2022lowresource}.

\setlength{\tabcolsep}{2.1pt}
\begin{table}[t]
\fontsize{9}{11}\selectfont
\begin{tabular}{lccccccc}
\toprule
 & \multicolumn{2}{c}{Style} & \multicolumn{2}{c}{UAR} & & \\
\cmidrule(lr){2-3} \cmidrule(lr){4-5}
Method & Conf. & Joint & Conf. & Joint & Sim & Fluency \\
\midrule
            \textsc{Para} & 0.42 & 0.335 & 0.26 & 0.202 & 0.64 & \textbf{0.85} \\
            \textsc{BERT} &  0.31 & 0.076 & 0.30 & 0.061 & 0.13 & 0.35 \\
            \textsc{LING} &  0.44 & 0.334 & 0.23 & 0.177 & \textbf{0.82} & 0.58 \\
STRAP$_{\textit{fine-tuned}}$ &  0.47 & 0.344 & 0.32 & \underline{0.218} & 0.54 & 0.83 \\
\midrule
ChatGPT-3.5 &  0.54 & 0.338 & \textbf{0.48} & \textbf{0.280} & 0.56 & 0.79 \\
\hline
\hline
PGuide ($\lambda=2.5 \mathrm{e}3$) &  \textbf{0.74} & 0.431 & \underline{0.36} & 0.209 & 0.42 & 0.64 \\
PGuide ($\lambda=1.5 \mathrm{e}3$) &  0.68 & \textbf{0.434} & 0.33 & 0.207 & 0.47 & 0.70 \\
PGuide ($\lambda=8 \mathrm{e}2$) &  0.64 & 0.426 & 0.33 & 0.217 & 0.50 & 0.74 \\
PGuide ($\lambda=2 \mathrm{e}2$) &  0.50 & 0.353 & 0.29 & 0.204 & 0.52 & 0.78 \\
\bottomrule
\end{tabular}
\caption{Evaluation metrics for authorship style transfer. We evaluate using two authorship representations: \textit{Style} \cite{wegmann-etal-2022-author} and \textit{UAR} \cite{rivera-soto-etal-2021-learning}. For each metric, we bold the strongest approach, and underline the most performant non-LLM method.}
\label{table:author_eval}
\end{table}
\begin{figure}[t]
\includegraphics[width=8cm]{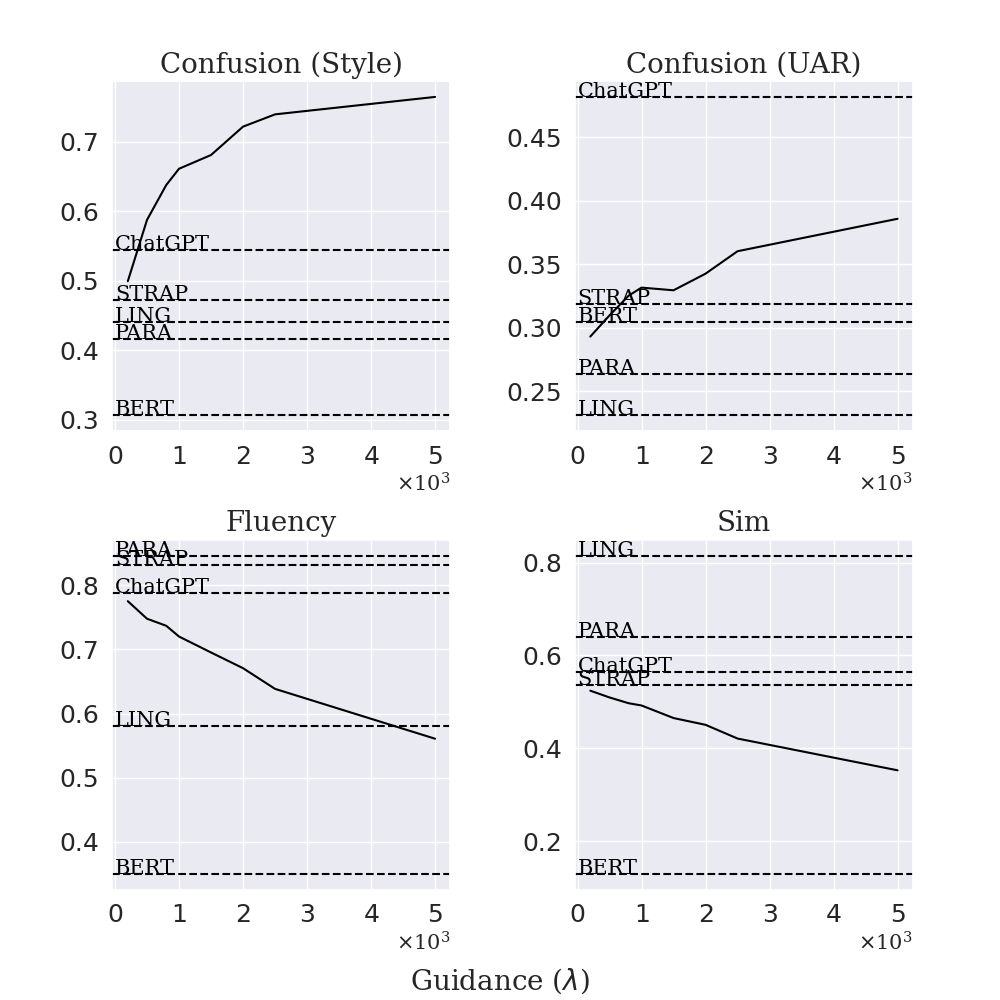}
\caption{As we increase the guidance hyperparameter $\lambda$, we steadily increase style transfer accuracy (\textit{Confusion}), at the cost of semantic consistency (\textit{Sim}) and \textit{Fluency}.}
\label{fig:tradeoff}
\end{figure}

\subsection{Authorship Style Transfer}
Table \ref{table:author_eval} presents our results on the challenging task of low-resource authorship style transfer. When evaluated with the \textit{Style} embedding space, three of the four \textsc{ParaGuide} configurations outperform every single baseline (including ChatGPT-3.5) on \textit{Joint} and \textit{Confusion}. When we consider the holdout \textit{UAR} embedding space, however, ChatGPT-3.5, which notably uses $400$x more parameters than \textsc{ParaGuide}, outperforms the other approaches.
Considering only non-LLM methods, \textsc{ParaGuide} outperforms all baselines on \textit{UAR Confusion}, but is very narrowly outperformed by \textsc{Strap} on \textit{UAR Joint}. This can be attributed, however, to \textsc{Strap}'s higher \textit{Fluency} score, which was a metric that was not predictive of human ratings on the formality task. Additionally, in contrast to \textsc{ParaGuide}'s plug-and-play approach, the \textsc{Strap} implementation involves $110$ separate models, each with $800$ million parameters, fine-tuned for every author.

\vspace{18pt}

\subsection{Style Transfer vs. Similarity and Fluency}

Beyond showcasing \textsc{ParaGuide}'s strong performance, our automatic evaluations in Tables \ref{table:formal_auto}, \ref{table:sentiment_auto}, and \ref{table:author_eval} demonstrate control over the trade-off between transfer accuracy versus semantic consistency and fluency, via the $\lambda$ hyperparameter. 
We additionally visualize the affect of varying $\lambda$ on authorship style transfer in Figure \ref{fig:tradeoff}.
 When $\lambda$ is small, the paraphrase-conditioned diffusion model reconstructs a more semantically faithful, fluent output. However, we can increase $\lambda$ to improve \textit{Confusion} scores, at the cost of semantic consistency and fluency. At the lowest setting, \textsc{ParaGuide}'s \textit{Fluency} and \textit{Similarity} score are similar to those of ChatGPT-3.5 ($0.78$ vs $0.79$ and $0.52$ vs $0.56$).

\section{Conclusion and Future Work}

We introduce \textsc{ParaGuide}, a diffusion-based framework for unsupervised textual style transfer. The approach harnesses the controllability of text diffusion, alongside the availability of off-the-shelf text classifiers and stylistic embedders, to competitively perform both authorship and attribute transfer, without ever having to retrain style-specific pipelines.

Our work demonstrates the potential of diffusion for text generation, a landscape currently dominated by large, auto-regressive language models. We are particularly excited about pursuing work that explores scaling diffusion models, better adapting them to the text domain, and the ways that these non-autoregressive methods can work alongside and complement current state-of-the-art approaches.

\section*{Ethical Statement}
\textsc{ParaGuide} presents an effective diffusion-based framework for style-transfer that uses fewer parameters than other state-of-the-art methods, can be fine-tuned on a single GPU, and avoids having to retrain models for new target styles. 
As a result, the approach could broaden the accessibility of controllable text generation and empower individuals with fewer resources to better personalize systems to their needs. At the same time, we recognize that text generation approaches like ours have the potential to be leveraged by malicious actors for impersonation and persuasion.

\section*{Acknowledgements}
We would like to thank Xiaochuang Han, Raghav Singhal, Amith Ananthram, Debasmita Bhattacharya, Nicholas Deas, Maximillian Chen, and Smaranda Muresan for their invaluable discussions and thoughtful feedback, which helped shape the direction of this work. Additionally, we would like to extend our gratitude to
Samir Gadre, Fei-Tzin Lee, and Matthew Toles for their support on human evaluations, and our anonymous AAAI reviewers for their comments.

This research is supported in part by the Office of the Director of National Intelligence (ODNI), Intelligence Advanced Research Projects Activity (IARPA), via the HIATUS Program contract \#2022-22072200005. The views and conclusions contained herein are those of the authors and should not be interpreted as necessarily representing the official policies, either expressed or implied, of ODNI, IARPA, or the U.S. Government. The U.S. Government is authorized to reproduce and distribute reprints for governmental purposes notwithstanding any copyright annotation therein.

\appendix

\vspace{.2em}

\section{Additional Model Details}

\subsection{Paraphrase Generation}
The \textsc{Paraguide}, \textsc{Strap}, and \textsc{Para} approaches require paraphrases of input text. To generate these paraphrases, we use a publicly available Pegasus-based \cite{zhang2020pegasus} model fine-tuned for paraphrase generation.\footnote{\url{https://huggingface.co/tuner007/pegasus_paraphrase}}
Given the importance of diverse paraphrases \cite{krishna2020reformulating}, we perform nucleus sampling \cite{holtzman2020curious} with $\tau=1.5$ and $p=0.80$. We use the same sampling procedure to generate paraphrases at both training and inference time. 

\subsection{ParaGuide}

\subsubsection{Architecture}

We started with the publicly available SSD-LM RoBERTa-large checkpoint,\footnote{https://huggingface.co/xhan77/ssdlm} and made several modifications. First, we removed the \textit{embedding sum} layer, so that the model accepts noised word embedding latent representations. We then modify the architecture to take two fixed-sized inputs: the embedded paraphrase and the noised embedding representation. Both of these inputs are padded to $50$ tokens. 

\subsubsection{Training}

We train our paraphrase-conditioned model for $500$K steps with our revised noise schedule on our synthetic dataset of Reddit paraphrases, with $T=5000$ before fine-tuning on the Enron Corpus, with $T=200$.  We use a single NVIDIA A100 with a batch size of $128$ and learning rate of $5\mathrm{e}{-6}$, and selected the checkpoint with the lowest validation loss on non-holdout author data. 

\subsubsection{Inference}

At inference time, we perform nucleus sampling \cite{holtzman2020curious} at each diffusion timestep with $p=0.80$. We use $k=3$ optimization steps with our guidance models, with temperature $\tau=3$. We experimented with a range of $\lambda$ values on our validation set, and selected the final values to showcase different balances of transfer accuracy vs semantic consistency/fluency.

After observing that both \textsc{ParaGuide} and our \textsc{M\&M} baselines could occasionally return empty strings, we modified the inference procedure for both approaches to run a secondary inference if this occurred. Given that diffusion models have been shown to generate diverse text \cite{han2023ssdlm, yuan2023seqdiffuseq}, we expect a stronger inference procedure for our model would be to sample multiple outputs, and then select the best based on guidance models, but we leave this to future work. 

\subsubsection{Guidance Models}

We use an existing formality classifier\footnote{\url{https://huggingface.co/cointegrated/roberta-base-formality}} as our internal formality guidance model. For sentiment transfer, we employ an existing model fine-tuned on Twitter data \cite{barbieri-etal-2020-tweeteval}.\footnote{\url{https://huggingface.co/cardiffnlp/twitter-roberta-base-sentiment}}
For authorship transfer, we compute stylistic distance with Style Embeddings \cite{wegmann-etal-2022-author}.

\subsection{Baselines}

For our \textsc{Strap} baseline, we fine-tune a T5-large model \cite{raffel2020exploring} on the Reddit paraphrase dataset for $250$K steps with a batch size of $32$, gradient accumulation of $4$, and learning rate of $1\mathrm{e}{-5}$. We continue training on the non-holdout author Enron paraphrases with a learning rate of $1\mathrm{e}{-5}$ and selected the checkpoint with the lowest validation loss. For attribute transfer, we fine-tune four target-style specific models (formal, informal, positive sentiment, negative sentiment) on the holdout author data labeled as such by the external classifier. For authorship transfer, we fine-tune this base model for each of the $110$ holdout authors on $60\%$ of their data. For all \textsc{Strap} experiments, we fine-tune with a real batch size of $64$ and learning rate of $1\mathrm{e}{-4}$. At inference time, we perform greedy decoding.

For our \textsc{M\&M} baselines, we use a pretrained RoBERTa-large \cite{liu2019roberta} model as the masked language model, and the hyperparameter configurations specified in \citet{mireshghallah2022mix}, for sentiment and formality transfer. For all experiments, we set the number of samples per input to be $3$. For our $enron$ configuration, we fine-tuned on the non-holdout author data for approximately $50$k steps with a learning rate of $1\mathrm{e}{-5}$ and batch size of $128$. 
\textsc{M\&M} inference speeds were approximately $10$x slower than \textsc{ParaGuide}'s, which limited our ability to run extensive hyperparameter tuning. 

We directly adapted the \textsc{BERT} and \textsc{Ling} baselines from \citet{patel2022lowresource}. For \textsc{Para}, we use the initial paraphrases generated by the autoregressive paraphraser. For our ChatGPT baseline, we prompt ChatGPT-3.5 (`gpt-3.5-turbo')   with up to $16$ examples in a target author's style, followed by ``Can you rewrite the following email to make it look like the above author's style:" and the source email. We use the default decoding parameters, with a temperature and $p$ of $1.0$.

\section{Additional Dataset Details}

\subsection{Preprocessing (Enron)}

 We preprocessed the Enron Emails Corpus \cite{Klimt2004TheEC} by:
 \begin{itemize}
     \item Removing duplicates.
     \item Filtering out email threads.
     \item Dropping emails longer than $50$ RoBERTa tokens \cite{liu2019roberta}.
     \item Dropping all email addresses with fewer than $10$ messages.
 \end{itemize}
 This results in a cleaned dataset of $89917$ emails, sent from $1100$ addresses.
 
 We randomly select $110$ authors (10\%) for a holdout set. We manually reviewed these assignments to ensure that emails by the same apparent sender but from different
addresses (i.e, personal versus business) are assigned to the
same shard. We split the non-holdout author data into train/validation/test ($0.8, 0.1, 0.1$). We use the non-holdout author training data to train \textsc{ParaGuide} and our baselines.

\subsection{Preprocessing (Reddit)}

We select up to $10$ comments from $400$K users in the MUD Reddit corpus \cite{andrews2019learning, khan2021deep}. 
For each comment, we sample $1$ sentence $70\%$ of the time, $2$ consecutive sentences $20\%$ of the time, and $3$ sentences $10\%$ of the time. We skip extractions that are longer than $200$ characters, and additionally filter texts that are longer than $50$ RoBERTa tokens. This resulted in approximately $390$K samples. We split the Reddit users into train/validation/test groups ($0.9, 0.05, 0,05$).

\subsection{Attribute Transfer Data}

For attribute style transfer, we use our external classifiers for formality and sentiment to classify each email in the holodut author set. 
For formality, we use a classifier\footnote{\url{https://huggingface.co/s-nlp/xlmr_formality_classifier}} trained on the XFORMAL corpus \cite{briakou2021xformal}. For sentiment classification, we use an existing model\footnote{\url{https://huggingface.co/siebert/sentiment-roberta-large-english}} fine-tuned on $15$ diverse datasets \cite{hartmann2023}.

This results in:
\begin{itemize}
    \item $9317$ formal emails.
    \item $3249$ informal emails.
    \item $8473$ positive sentiment emails.
    \item $4000$ negative sentiment emails.
\end{itemize}

We then split this data into train/validation/test shards ($0.7, 0.15, 0.15$).  We perform our final evaluation using $500$ test source emails for each label, with the exception of informality, which only has $486$ test samples.

\subsection{Authorship Transfer Data}

For authorship style transfer, we divided the holdout emails for each author into train/validation/test shards ($0.60, 0.20, 0.20$). To generate our results, we select up to $5$ test-set emails per source author, and then transfer each of these to $5$ other randomly selected holdout authors. This results in $1915$ transfer pairs.

\section{Human Evaluation}

For our human evaluation of formality transfer, we selected the approaches with the highest automatic \textit{Joint} scores. We then randomly selected $100$ formal and $100$ informal sample texts from our Enron test set, along with the corresponding output from each model.

In light of recent concerns about crowdworkers relying on large language models \cite{veselovsky2023artificial}, we selected our annotators from within our department, choosing members who were not directly involved with the project. All annotators are native English speakers.

We provided these annotators with the original reference inputs and corresponding candidate outputs, and asked them to rate (\{0, 1\}) the \textit{Similarity}, \textit{Fluency}/Well-formedness, and \textit{Formality} of these candidates. Their instructions were as follows:

\textit{Each of you has been assigned a series of very short emails to review.  Each example consists of a reference and output text.  You are asked to evaluate the output text across three criteria:}
\begin{enumerate}
    \item \textit{Similarity to the reference. Do the output and reference have a similar meaning?} ($0$=No, $1$=Yes)
    \item \textit{Well-formedness. Does the output look like a reasonable email? Is it coherent?} ($0$=Badly-Formed, $1$=Well-formed)
    \item \textit{Formality. Does the output text sound formal or informal?} ($0$=Informal, $1$=Formal)
\end{enumerate}

We collected three annotations per example and use the majority vote to determine labels. We then evaluated \textit{Accuracy} by comparing the target style to the \textit{Formality} label. Additionally, we compute \textit{Joint} for each sample by combining the other metrics ($\textit{Accuracy} \times  \textit{Similarity} \times \textit{Fluency}$).

We applied Krippendorff's $\alpha$ to evaluate inter-annotator aggrement for each label:

    \vspace{5pt}
\begin{center}
    \begin{tabular}{|c|c|}
    
        \hline
        \textbf{Human Label} & \textbf{Krippendorff} \\ \hline
        Similarity & $0.48$ \\ \hline
        Fluency &  $0.63$ \\ \hline
        Formality & $0.13$ \\ \hline
    \end{tabular}
   \vspace{5pt}
   \end{center}

Additionally, we used $z$-tests to determine the significance of \textsc{ParaGuide} ($\lambda=2 \mathrm{e}2$)'s improvement relative to the next best approach, M\&M (Hamming).

\vspace{5pt}
\begin{center}
    \begin{tabular}{|c|c|c|}
    
        \hline
        \textbf{Metric} & $z$ & $p$ \\ \hline
        Joint & 3.10 & 0.00098 \\ \hline
        
         Accuracy
& 3.57 & 0.00018 \\ \hline
Similarity & 1.71 & 0.044 \\ \hline
Well-formedness & 4.88 & 5.36e-07 \\ \hline

    \end{tabular}
    \end{center}
   \vspace{5pt}

All results are significant with $p=0.05$.

\section{Reference Outputs}

Tables 
\ref{table:informal_examples},
\ref{table:formal_examples},
\ref{table:negative_examples},
and \ref{table:positive_examples}
contain representative examples of model output for each attribute transfer task. Tables \ref{table:authorship_example_1} and \ref{table:authorship_example_2} present authorship style transfer examples. All results are not cherry-picked. While the Enron dataset is publicly available, we do, however, exclude examples that include significant amounts of personal information. Additionally, we truncate long texts.

\setlength{\tabcolsep}{3pt}.
\begin{table*}[t]
\centering
\begin{tabular}{llccc}
\toprule
\textbf{Original Text} & Thursday afternoon would be best for me. Enron North America Corp. \\
ParaGuide ($\lambda=1e4$) & Thu afternoon works best with me.. \\
ParaGuide ($\lambda=5e3$) & Thursday afternoon would be good for mine \\
ParaGuide ($\lambda=1e3$) & Thursday afternoon works best for me?? \\
ParaGuide ($\lambda=5e2$) & Thursday afternoon would work best for me... \\
ParaGuide ($\lambda=2e2$) & Thu afternoon would be best with me!Says \\
STRAP$_{\textit{fine-tuned}}$ & .. I'm going to be indoors on Thursday afternoon. I'm going to be indoors on Thu[...] \\
M\&M (Hamming) & That would be best for me, Enron North America Corp \\
\midrule
\textbf{Original Text} & Well, don't get too excited until you read it. There's lots of stuff I just had[...] \\
ParaGuide ($\lambda=1e4$) & i just put place holders on but we will get it all filled out, don't get excited[...] \\
ParaGuide ($\lambda=5e3$) & ont get excited until you read it.... I just put place holders on but we'll get[...] \\
ParaGuide ($\lambda=1e3$) & don't get excited until you read it, I just put place holders on but we'll get i[...] \\
ParaGuide ($\lambda=5e2$) & don't get excited until you read it - I just put place holders on but we will ge[...] \\
ParaGuide ($\lambda=2e2$) & don't get excited until you read it, I just put place holders on but we will get[...] \\
STRAP$_{\textit{fine-tuned}}$ & I haven't done anything yet, I will get it all filled in... on the table, I just[...] \\
M\&M (Hamming) & don't get too excited until you read it :) There's lots of stuff I just had to p[...] \\
\midrule
\textbf{Original Text} & Jeff, I think this expands your scope. (Which is, of course, a good thing as lon[...] \\
ParaGuide ($\lambda=1e4$) & Thanks Sue - i feel like this expands some scope \& good stuff! \\
ParaGuide ($\lambda=5e3$) & Hi Sue...I think this really expands your scope...Thanks > \\
ParaGuide ($\lambda=1e3$) & good work, doc - i feel like it expands your scope. Sue \\
ParaGuide ($\lambda=5e2$) & thanks Sue - i think this expands your scope its a good thing \\
ParaGuide ($\lambda=2e2$) & sue- i think it expands your scope and generally a good thing. \\
STRAP$_{\textit{fine-tuned}}$ & , I think that this is a good thing. I think that this expands your scope and is[...] \\
M\&M (Hamming) & I think this expands your scope tremendously (Which is, of course, a good thing[...] \\
\midrule
\textbf{Original Text} & The new market readiness plan appears to target March 1, 2002 as the market open[...] \\
ParaGuide ($\lambda=1e4$) & On the GANTT chart tabs end Mar 01,,2002,, ds look like the market opening date[...] \\
ParaGuide ($\lambda=5e3$) & looks like the bars go off GANTT chart March 1 in NEW MARK READY OPEN PLAN. \\
ParaGuide ($\lambda=1e3$) & got all - the bars in any GANTT chart end March 1, 2002- that appears as NEW mar[...] \\
ParaGuide ($\lambda=5e2$) & i think the expiration bars in GDT chart - March 01 - 2002.... it looks like mar[...] \\
ParaGuide ($\lambda=2e2$) & ay - note bars in the GANTTUM chart end on march 1,2002.. so that is the market[...] \\
STRAP$_{\textit{fine-tuned}}$ & ... The bars end on March 1, 2002, so that appears in the new market readiness p[...] \\
M\&M (Hamming) & new market readiness plan appears to target March 1, 2002 as the market opening[...] \\
\midrule
\textbf{Original Text} & He's re-trading on the retainage. Let's discuss. Kay \\
ParaGuide ($\lambda=1e4$) & he's re-trading on the retainage...let's discuss \\
ParaGuide ($\lambda=5e3$) & hes retrading the reiterate -- lets talk \\
ParaGuide ($\lambda=1e3$) & let me discuss he's reading on the retainage \\
ParaGuide ($\lambda=5e2$) & Yeah he's re-trading on the retainage? Lets discuss!! \\
ParaGuide ($\lambda=2e2$) & let me discuss... he's retrading on the retainage \\
STRAP$_{\textit{fine-tuned}}$ & the retainage. the retainage. he's re-trading on the retainage.re-trading on the[...] \\
M\&M (Hamming) & re-trading on the retainage. Let's discuss. \\
\midrule
\textbf{Original Text} & Joanne: please change the attached give-up in Par. 10 for provide for New York ([...] \\
ParaGuide ($\lambda=1e4$) & OK I print three signatures off today and amend give up Par.10 NOT NEW \\
ParaGuide ($\lambda=5e3$) & lets print three signatures - change the give-up in Par.10 and reflect New York[...] \\
ParaGuide ($\lambda=1e3$) & ok just need to print three signatures \& change give up on Par. 10 acknowledging[...] \\
ParaGuide ($\lambda=5e2$) & please print 3 signatures and change give-up in Par.10 so to New York instead of[...] \\
ParaGuide ($\lambda=2e2$) & please print three signatures also and change the give ups in Par. 10 to reflect[...] \\
STRAP$_{\textit{fine-tuned}}$ & . I need to print three signatures. Thanks. I need to print three signatures. I[...] \\
M\&M (Hamming) & PLEASE change the overall give-up in Par. 10 for ONLY for New York (not NY). The[...] \\
\bottomrule
\end{tabular}
\caption{Formal $\rightarrow$ Informal Examples.}
\label{table:informal_examples}
\end{table*}

\begin{table*}[t]
\centering
\begin{tabular}{llccc}
\toprule
\textbf{Original Text} & need defn of above \\
ParaGuide ($\lambda=1e4$) & Defect above \\
ParaGuide ($\lambda=5e3$) & DefRN requires above \\
ParaGuide ($\lambda=1e3$) & Defol Below \\
ParaGuide ($\lambda=5e2$) & DefN of above required \\
ParaGuide ($\lambda=2e2$) & Requires defn of above \\
STRAP$_{\textit{fine-tuned}}$ & . There is a need of defn of above.... above... of above. of above. of above. of[...] \\
M\&M (Hamming) & of \\
\midrule
\textbf{Original Text} & - PJM.xls \\
ParaGuide ($\lambda=1e4$) & JPJM."xls \\
ParaGuide ($\lambda=5e3$) & MQWM.Xls \\
ParaGuide ($\lambda=1e3$) & PPGM.xls \\
ParaGuide ($\lambda=5e2$) & JPM.xls \\
ParaGuide ($\lambda=2e2$) & JPM.xls \\
STRAP$_{\textit{fine-tuned}}$ & for PJM.xls..xls..xls.xls.xls.M.xls. The PJM.xls is an acronym for PJM... \\
M\&M (Hamming) & PJM. \\
\midrule
\textbf{Original Text} & yes, yes, and yes. You're on a roll.... \\
ParaGuide ($\lambda=1e4$) & Yes. \\
ParaGuide ($\lambda=5e3$) & Absolutely rolling \\
ParaGuide ($\lambda=1e3$) & Yes \\
ParaGuide ($\lambda=5e2$) & Yes and Jeffrey \\
ParaGuide ($\lambda=2e2$) & Yes \\
STRAP$_{\textit{fine-tuned}}$ & and yes, and yes, and yes. Yes, and yes. and yes. Yes, and yes. Yes. Yes. Yes. Y[...] \\
M\&M (Hamming) & yes, and yes. You're on a \\
\midrule
\textbf{Original Text} & Who's your buddy???????????????????????????? A little hint wouldn't kill you, wo[...] \\
ParaGuide ($\lambda=1e4$) & A little hint wouldn't kill you... however I will disclose everything \\
ParaGuide ($\lambda=5e3$) & Minor hints platnt kill you.--I hereby disclose. \\
ParaGuide ($\lambda=1e3$) & Little teasing Should not kill you. Lets disclose everything \\
ParaGuide ($\lambda=5e2$) & A little hint wouldn't kill me. I will share everything. Bx \\
ParaGuide ($\lambda=2e2$) & I will tell you all details, if a little hint would kill you. \\
STRAP$_{\textit{fine-tuned}}$ & you know everything. I will let you know everything. I will let you know everyth[...] \\
M\&M (Hamming) & your favorite???????????????????????????? A little hint wouldn't kill you, would[...] \\
\midrule
\textbf{Original Text} & fyi \\
ParaGuide ($\lambda=1e4$) & wp \\
ParaGuide ($\lambda=5e3$) & Yu \\
ParaGuide ($\lambda=1e3$) & TP \\
ParaGuide ($\lambda=5e2$) & really quite comprehend you \\
ParaGuide ($\lambda=2e2$) & I'm not sure what you are referring to. \\
STRAP$_{\textit{fine-tuned}}$ & . I'm not sure what you're talking about. I'm not sure what you're talking about[...] \\
M\&M (Hamming) &  \\
\midrule
\textbf{Original Text} & FYI \\
ParaGuide ($\lambda=1e4$) & True. \\
ParaGuide ($\lambda=5e3$) & FY., \\
ParaGuide ($\lambda=1e3$) & True true \\
ParaGuide ($\lambda=5e2$) & True \\
ParaGuide ($\lambda=2e2$) & True \\
STRAP$_{\textit{fine-tuned}}$ & . It's true. It's true. It's true. It's true. It's true...... It's true... It's[...] \\
M\&M (Hamming) &  \\
\midrule
\bottomrule
\end{tabular}
\caption{Infomal $\rightarrow$ Formal Examples.}
\label{table:formal_examples}
\end{table*}

\begin{table*}[t]
\centering
\begin{tabular}{llccc}
\toprule
\textbf{Original Text} & Thanks, Tana! -MP \\
ParaGuide ($\lambda=1e4$) & FU DIS!!!!!!!!!!!!!!!! \\
ParaGuide ($\lambda=5e3$) & thanks fuckinea!!!!!!!! \\
ParaGuide ($\lambda=1e3$) & Thanks, tana? \\
ParaGuide ($\lambda=5e2$) & Tana????? \\
ParaGuide ($\lambda=2e2$) & Seeing you Tana????? \\
STRAP$_{\textit{fine-tuned}}$ & ! I appreciate your kind words. You are so kind, Tana! I appreciate your kind wo[...] \\
M\&M (Hamming) & Tana … \\
\midrule
\textbf{Original Text} & when? \\
ParaGuide ($\lambda=1e4$) & ????? \\
ParaGuide ($\lambda=5e3$) & abuse????? \\
ParaGuide ($\lambda=1e3$) & ????? when \\
ParaGuide ($\lambda=5e2$) & when????? \\
ParaGuide ($\lambda=2e2$) & when????? \\
STRAP$_{\textit{fine-tuned}}$ & When? \\
M\&M (Hamming) &  \\
\midrule
\textbf{Original Text} & I may be on the road tomorrow (won't know until later), so something early after[...] \\
ParaGuide ($\lambda=1e4$) & Early afternoon early??sorry screw thru suck!!!! \\
ParaGuide ($\lambda=5e3$) & something shitty sucks Chaturs hate thinking road tomorrow. \\
ParaGuide ($\lambda=1e3$) & too awful!!!!! I'm driving tomorrow. Probably would prefer later afternoon Frida[...] \\
ParaGuide ($\lambda=5e2$) & Too busy!!!! I'd prefer something early afternoon Friday (may actually be drivin[...] \\
ParaGuide ($\lambda=2e2$) & because dude knows.......i may be traveling early tomorrow....I'd prefer somethi[...] \\
STRAP$_{\textit{fine-tuned}}$ & . I would prefer something early morning Friday because I would be on the road t[...] \\
M\&M (Hamming) & may be on the road tomorrow (won't know until later), so retiring early afternoo[...] \\
\midrule
\textbf{Original Text} & Thanks for the catch! Sara \\
ParaGuide ($\lambda=1e4$) & Congratulations Sierra...... \\
ParaGuide ($\lambda=5e3$) & Sorry Sheila!!!!!!!! \\
ParaGuide ($\lambda=1e3$) & Susan Sheila catching?????!!!!!!!! \\
ParaGuide ($\lambda=5e2$) & Sorry catch...... Sue \\
ParaGuide ($\lambda=2e2$) & hello eh the catch!!!!!!!! \\
STRAP$_{\textit{fine-tuned}}$ & I'm glad you've fixed it! Thanks, Thanks, I'll fix it! Thanks, Thanks, Sara! Tha[...] \\
M\&M (Hamming) & me the catch! \\
\midrule
\textbf{Original Text} & Section B2 contains the same argument that we made to FERC in GridFlorida (menti[...] \\
ParaGuide ($\lambda=1e4$) & same argument GridFlorida210!!!!! due fuck Monday!!!!! \\
ParaGuide ($\lambda=5e3$) & Report spawned stupid OUR VERY same argument int=/2 over GridFlorida \\
ParaGuide ($\lambda=1e3$) & FYIs????? Due 61 while ours made....Same argument against section B2 in GridFlor[...] \\
ParaGuide ($\lambda=5e2$) & It's due monday morning, we make the same arguments in SECTION B02 in GridFlorid[...] \\
ParaGuide ($\lambda=2e2$) & ASI is due on Monday......We made the same argument in Section B02 in GridFlorid[...] \\
STRAP$_{\textit{fine-tuned}}$ & .. Thanks for the link. I'll look into it... We made the argument in Section B2[...] \\
M\&M (Hamming) & B2 contains the same complaints that we made to FERC in GridFlorida (mentions Ca[...] \\
\midrule
\textbf{Original Text} & also 519128 and 519230 \\
ParaGuide ($\lambda=1e4$) & 548129 \\
ParaGuide ($\lambda=5e3$) & 54819159 \\
ParaGuide ($\lambda=1e3$) & both 55019129 55039 \\
ParaGuide ($\lambda=5e2$) & Also 55019127 & 55031 \\
ParaGuide ($\lambda=2e2$) & both 519128 & 52530 \\
STRAP$_{\textit{fine-tuned}}$ & . 51912 and 522530 were also included.... 519124 was also included..530. 519124[...] \\
M\&M (Hamming) & 519128 not \\

\bottomrule
\end{tabular}
\caption{Positive $\rightarrow$ Negative Examples.}
\label{table:negative_examples}
\end{table*}

\begin{table*}[t]
\centering
\begin{tabular}{llccc}
\toprule

\textbf{Original Text} & Louise, Do we continue to wear the risk/reward and carrying charges on the Mitsu[...] \\
ParaGuide ($\lambda=1e4$) & do your support welcome our risk/reward plus continue carrying Pere Mitsubishi c[...] \\
ParaGuide ($\lambda=5e3$) & May we continue delivering on risk/reward for our charges E Mitsubishis? Awesome[...] \\
ParaGuide ($\lambda=1e3$) & We continue wearing the value/reward and charging charges on our M Mitsubishis?[...] \\
ParaGuide ($\lambda=5e2$) & So I welcome the Risk/reward and continue carrying charges on the Mitsubishis. T[...] \\
ParaGuide ($\lambda=2e2$) & Thanks Mike We should continue wearing our risk/reward and continue carrying cha[...] \\
STRAP$_{\textit{fine-tuned}}$ & .. Chris Calger asked if we should continue wearing the risk/reward and carrying[...] \\
M\&M (Hamming) & Do we continue to wear the risk/reward and carrying charges on the Mitsubishis?[...] \\
\midrule
\textbf{Original Text} & I'm swamped at the moment and I'll be in South America next week. I then have va[...] \\
ParaGuide ($\lambda=1e4$) & Hey Mike, I'm going in Chile next week and I am grateful for your help!!!! Just[...] \\
ParaGuide ($\lambda=5e3$) & Awesome!! I'm going to Chile next week. Just let me know!! thanks \\
ParaGuide ($\lambda=1e3$) & Just let me know this will help. I am planning to come to South America next wee[...] \\
ParaGuide ($\lambda=5e2$) & I am on my way to South America next week. I am definitely around and could help[...] \\
ParaGuide ($\lambda=2e2$) & I'm going over to South America this summer starting next week plus expecting so[...] \\
STRAP$_{\textit{fine-tuned}}$ & . I need help. I'm going to South America next week and I need help. I need help[...] \\
M\&M (Hamming) & swamped at the moment and I'll be in South America next week. I then have vacati[...] \\
\midrule
\textbf{Original Text} & sorry...when did you get out of here yesterday? \\
ParaGuide ($\lambda=1e4$) & its nice great. did you get out here yesterday! \\
ParaGuide ($\lambda=5e3$) & Thanks! Did your guys acron out yesterday!! Congratulations! \\
ParaGuide ($\lambda=1e3$) & Thanks! Did you get out of here yesterday!! Thanks! \\
ParaGuide ($\lambda=5e2$) & Thanks! Did you get out of here yesterday!! Thanks!!! \\
ParaGuide ($\lambda=2e2$) & did yo get out of here yesterday! we are good o. D \\
STRAP$_{\textit{fine-tuned}}$ & did you get out of here yesterday? I'm sorry, but did you get out of here yester[...] \\
M\&M (Hamming) & did you get out of here \\
\midrule
\textbf{Original Text} & PLEASE DISREGARD THIS INFO. WE ARE NO LONGER DOING THE LENDING DEAL [...] \\
ParaGuide ($\lambda=1e4$) & Love Bernie VirFranc EPS We supportForward Garage \\
ParaGuide ($\lambda=5e3$) & ourperforming conquering Beaver 38993 \\
ParaGuide ($\lambda=1e3$) & Our keep working In our loan deal closed. \\
ParaGuide ($\lambda=5e2$) & We are bank executed our new deal thru Friday. \\
ParaGuide ($\lambda=2e2$) & This is the Loan deal for Friday and Saturday, we are definitely doing this week[...] \\
STRAP$_{\textit{fine-tuned}}$ & deal on Friday and Saturday.. We are not doing the loan deal on Friday and Satur[...] \\
M\&M (Hamming) & DISREGARD THIS POST. WE ARE NO LONGER DOING THE LENDING DEAL [...] \\
\midrule
\textbf{Original Text} & if I were a rich man.... or I'll call you later. \\
ParaGuide ($\lambda=1e4$) & Oh you are pretty incredibly rich! I'll call you later! Congratulations! \\
ParaGuide ($\lambda=5e3$) & Congratulations! My thoughts were totally rich man! I'll call you later! \\
ParaGuide ($\lambda=1e3$) & only I was sweetest! i will call you later \\
ParaGuide ($\lambda=5e2$) & If I was rich bought, I will call you later! Congratulations. \\
ParaGuide ($\lambda=2e2$) & if i was rich i'll call you later :) \\
STRAP$_{\textit{fine-tuned}}$ & I'll tell you later, if you're rich. I'd be rich. I'll tell you later, if I'm ri[...] \\
M\&M (Hamming) & I were a rich man.... or I'll call you \\
\midrule
\textbf{Original Text} & It is not intended to be a negotiated rate. We will add the other language when[...] \\
ParaGuide ($\lambda=1e4$) & that's negotiated rate love me!! Vince \\
ParaGuide ($\lambda=5e3$) & 801MM favorably really agreefully rate \\
ParaGuide ($\lambda=1e3$) & non bid \\
ParaGuide ($\lambda=5e2$) & doesn't negrated rate :) \\
ParaGuide ($\lambda=2e2$) & Means for point means being a negotiated rate! \\
STRAP$_{\textit{fine-tuned}}$ & a negotiated rate. It's a flat rate.. It's a guideline. It's a guideline. a fixe[...] \\
M\&M (Hamming) & is not intended to be a negotiated rate. We will add the other language when a d[...] \\

\bottomrule
\end{tabular}
\caption{Negative $\rightarrow$ Positive Examples.}
\label{table:positive_examples}
\end{table*}

\begin{table*}[t]
\centering
\begin{tabular}{llccc}
\toprule
\textbf{Target Author Examples} & \\
Some very nice, and very appropriate thoughts about John Ambler. Wade continues[...] \\
Here's an early draft. Note Cinergy sr. mgmt has not yet commented. No price, ei[...] \\
Here's the draft that Allegretti is sending. \\
here you go \\
Sounds pretty harmless to me. If you don't want to do it, let me know and I'll p[...] \\
... \\
\midrule
\end{tabular}
\begin{tabular}{llccc}
\textbf{Original Text} & Thanks. Comments below. Great input. \\
\midrule
PARA & I would like to thank you for the great input. \\
BERT & Thanks. Thanks fine. Encouraging price. \\
LING & Thanks. comments below. great input. \\
STRAP$_{\textit{fine-tuned}}$ & Thanks for the great input. Mark \\
ChatGPT & Here's my feedback. Great points. Mark \\
ParaGuide ($\lambda=200$) & Thanks for the great input. BT \\
ParaGuide ($\lambda=500$) & Thanks for the great input \\
ParaGuide ($\lambda=800$) & Thanks for the great input! \\
ParaGuide ($\lambda=1000$) & thanks for the great input. \\
ParaGuide ($\lambda=1500$) & Great input. Thanks! \\
ParaGuide ($\lambda=2000$) & Thanks for the great input \\
ParaGuide ($\lambda=2500$) & Thanks for the great input, BT \\
ParaGuide ($\lambda=5000$) & It's truly useful input. Thanks! Mark \\
\midrule
\bottomrule
\end{tabular}
\caption{Authorship Transfer Example $1$. In this example, as we increase $\lambda$, we can observe several changes, including the introduction of a contraction, which matches several sample texts. Additionally, like \textsc{Strap} and ChatGPT, with a larger $\lambda$, \textsc{ParaGuide} also splits the output into two sentences, and adds the `Mark' signature. While this is not visible in the truncated input, the target author signs their emails with `Mark'.}
\label{table:authorship_example_1}
\end{table*}

\begin{table*}[t]
\centering
\begin{tabular}{llccc}
\toprule
\textbf{Target Author Examples} & \\
scriptsnetscape.wbt "http://x12.enron.com/hottap.htm" \\
IT'S NOT JO JO - IT'S J-LO I THINK SHE WAS ON AT 8 HAS ANYONE JUMPED OUT THE WIN[...] \\
>> This is something you should see >> Puts things in perspective. >> >> http://[...] \\
so YOU watched it?? \\
i did hear that it was rather revealing \\
... \\
\midrule
\end{tabular}
\begin{tabular}{llccc}
\textbf{Original Text} & Then why are you dating a girl from LSU? \\
\midrule
PARA & Why do you want to be with a girl from LSU? \\
BERT & So how are YOU trying the perspective from LSU? \\
LING & then why am you dating a girl from lsu"" \\
STRAP$_{\textit{fine-tuned}}$ &  why are you dating a girl from LSU?  \\
ChatGPT & so YOU dating a girl from LSU?? \\
ParaGuide ($\lambda=200$) & why do you want to be with a girl from LSU? \\
ParaGuide ($\lambda=500$) & why do you want to be with a girl from LSU? \\
ParaGuide ($\lambda=800$) & why do you want to be with a girl from LSU? \\
ParaGuide ($\lambda=1000$) & why do you want to be with a girl from LSU \\
ParaGuide ($\lambda=1500$) & why do you want to be with a girl from LSU? \\
ParaGuide ($\lambda=2000$) & why do you need to be with a girl from LSU? \\
ParaGuide ($\lambda=2500$) & why do you need to be with a girl from LSU? \\
ParaGuide ($\lambda=5000$) & why do u need to be with some chicks from LSU?? \\
\midrule
\bottomrule
\end{tabular}
\caption{Authorship Transfer Example $2$. In this example, \textsc{ParaGuide }  lowercases the outputs, and uses more informal substitutions at $\lambda=5000$, including `you' $\rightarrow$ `u'.}
\label{table:authorship_example_2}
\end{table*}

\section{Resources}

The following sections detail the resources used in our experiments.

\subsection{Models}
\begin{itemize}
      \item SSD-LM \cite{han2023ssdlm}
      \item T5 (t5-Large) \cite{raffel2020exploring}
      \item ChatGPT (gpt3.5-turbo)
      \item Style Embeddings \cite{wegmann-etal-2022-author}
    \item Universal Authorship Representations \cite{rivera-soto-etal-2021-learning}
    \item RoBERTa (roberta-large) \cite{liu2019roberta}
    \item BERT \cite{devlin2019bert}
     \item Mutual Implication Score \cite{babakov-etal-2022-large}
      \item textattack/roberta-base-CoLA \cite{morris2020textattack,warstadt-etal-2019-neural}\footnote{https://huggingface.co/textattack/roberta-base-CoLA}
    \item cardiffnlp/twitter-roberta-base-sentiment \cite{barbieri-etal-2020-tweeteval}\footnote{\url{https://huggingface.co/cardiffnlp/twitter-roberta-base-sentiment}}
    \item siebert/sentiment-roberta-large-english \cite{hartmann2023}\footnote{\url{https://huggingface.co/siebert/sentiment-roberta-large-english}}
    \item cointegrated/roberta-base-formality\footnote{\url{https://huggingface.co/cointegrated/roberta-base-formality}}
    \item s\-nlp/xlmr\_formality\_classifier \cite{briakou2021xformal}\footnote{\url{https://huggingface.co/s-nlp/xlmr_formality_classifier}}
    \item tuner$007$/pegasus\_paraphrase \cite{zhang2020pegasus}\footnote{\url{https://huggingface.co/tuner007/pegasus_paraphrase}}

\end{itemize}

\subsection{Datasets}

\begin{itemize}
    \item Enron Email Corpus \cite{Klimt2004TheEC}
  \item Million User Dataset \cite{andrews2019learning, khan2021deep}
  \item WordNet \cite{miller-1994-wordnet}
  
\end{itemize}

\subsection{Software}
\begin{itemize}
    \item Transformers \cite{wolf2020huggingfaces}
    \item Datasets \cite{lhoest-etal-2021-datasets}
    \item Accelerate \cite{accelerate}
    \item Sentence-transformers 
    \cite{reimers2019sentencebert}
    \item spaCy  (en\_core\_web\_trf-3.3.0) \cite{ines_montani_2023_8225292} 
    \item NLTK \cite{nltk}
 \item pycontractions\footnote{\url{https://pypi.org/project/pycontractions}}
\item lemminflect\footnote{\url{https://pypi.org/project/lemminflect/}}
    \item Weights and Biases \cite{wandb}
    \item pySBD \cite{sadvilkar-neumann-2020-pysbd}
    \item statsmodels \cite{seabold2010statsmodels}
    \item NumPy \cite{harris2020array}
    \item Pandas \cite{reback2020pandas}
    \item PyTorch \cite{NEURIPS2019_9015}

\end{itemize}

\subsection{Compute}

We estimate the total compute budget for our experiments (model training and inference) as follows:
\begin{itemize}
   \item 1x NVIDIA A100 Tensor Core GPU / 100GB RAM / 4x CPU – 660 hours
\end{itemize}

\bigskip

\bibliography{aaai24}

\end{document}